\documentclass{article}
\usepackage[utf8]{inputenc}
\usepackage{abstract}

\usepackage{palatino}
\usepackage{amsfonts}
\usepackage{lipsum}
\usepackage{graphicx}
\usepackage{array}
\usepackage{xcolor}
\usepackage[backref=page,pagebackref=true,colorlinks,citecolor=citecolor,urlcolor=magenta,linkcolor=magenta]{hyperref}
\usepackage{url}
\usepackage{algorithmicx}
\usepackage{CJKutf8}
\usepackage{algorithm}
\usepackage{algpseudocode}

\usepackage[verbose=true,letterpaper]{geometry}
\newgeometry{
    textheight=9in,
    textwidth=6in,
    top=1in,
    headheight=14pt,
    headsep=25pt,
    footskip=30pt
}
\usepackage{fancyhdr}
\fancyheadoffset{0pt}
\usepackage{footmisc}
\renewcommand{\footnoterule}{\kern -3pt \hrule width 0.38\textwidth \kern 2.6pt}

\usepackage[square,numbers]{natbib}

\usepackage{longtable}
\usepackage{enumitem}
\usepackage{booktabs}
\usepackage{multicol}
\usepackage{multirow}
\usepackage{caption}
\usepackage{subcaption}
\usepackage{amsmath} 

\title{
\textbf{
Tiger200K: Manually Curated High Visual \\ Quality Video Dataset from UGC Platform}
}

\author{
\normalsize{}
Xianpan Zhou\hspace{3mm} 
\\
\vspace{7mm}
\normalsize{}
\vspace{-7mm}
}

\date{}
\begin{document}
\definecolor{citecolor}{HTML}{0071BC}

\maketitle

\begin{abstract}
\noindent
The recent surge in open-source text-to-video generation models has significantly energized the research community, yet their dependence on proprietary training datasets remains a key constraint. While existing open datasets like Koala-36M employ algorithmic filtering of web-scraped videos from early platforms, they still lack the quality required for fine-tuning advanced video generation models.
We present Tiger200K, a manually curated high visual quality video dataset sourced from User-Generated Content (UGC) platforms. By prioritizing visual fidelity and aesthetic quality, Tiger200K underscores the critical role of human expertise in data curation, and providing high-quality, temporally consistent video-text pairs for fine-tuning and optimizing video generation architectures through a simple but effective pipeline including shot boundary detection, OCR, border detecting, motion filter and fine bilingual caption.
The dataset will undergo ongoing expansion and be released as an open-source initiative to advance research and applications in video generative models. Project page: \url{https://tinytigerpan.github.io/tiger200k/}
\end{abstract}


\begin{figure*}[h!]
\centering
\includegraphics[width= 0.8\linewidth]{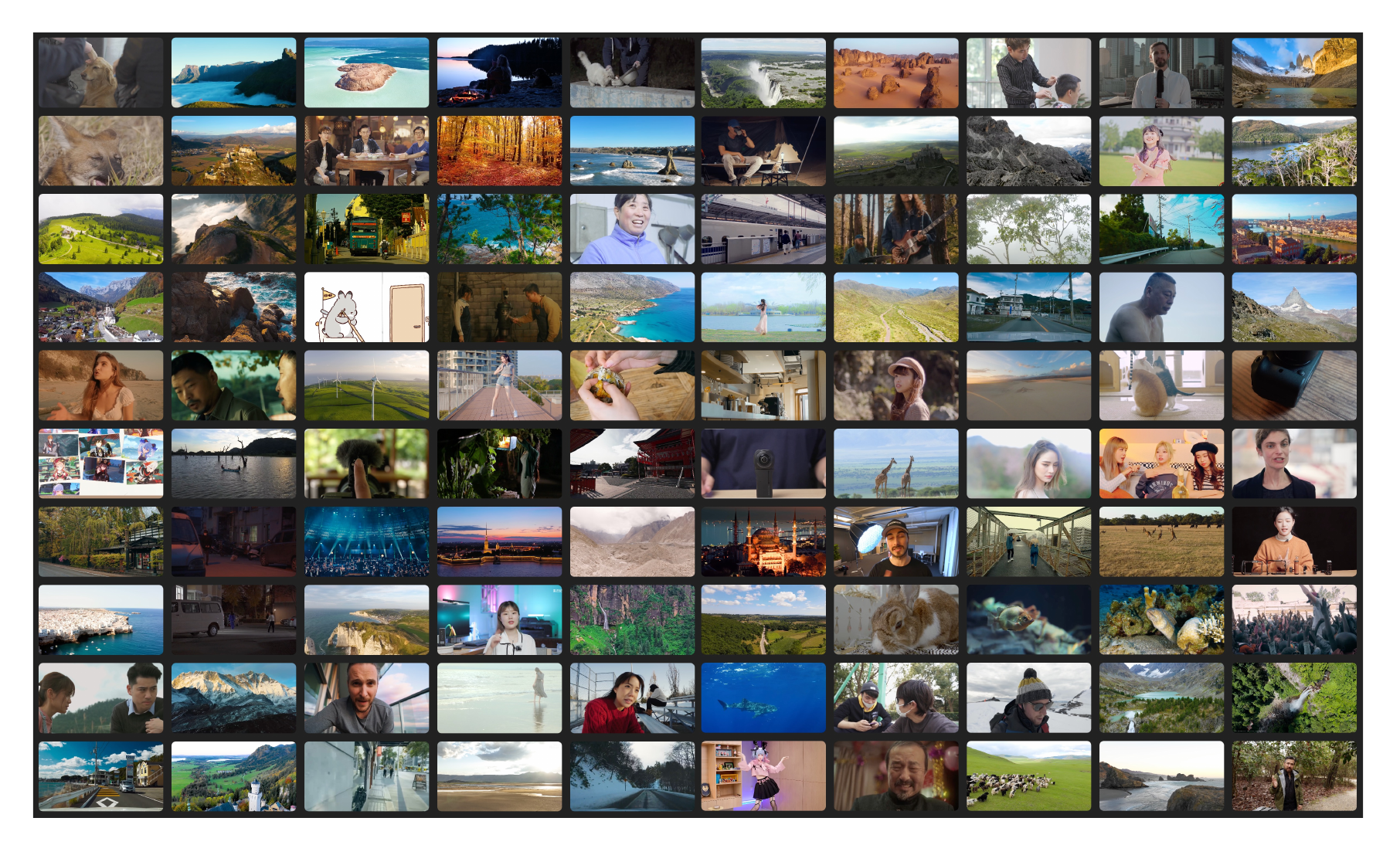}
   \caption{Visualization of randomly sampled clips in Tiger200k dataset. These clips demonstrate enhanced visual and aesthetic quality thanks to professional human curation in the data selection and review process and quality filtering data pipeline.}
\label{fig:visualization}
\end{figure*}

\section{Introduction}

Recent advances in generative artificial intelligence models have garnered significant research interest in computer vision, particularly in text-to-video generation tasks. Since OpenAI's introduction of Sora~\cite{SoraOpenAI}, the scaling up of large-scale datasets and models has led to the emergence of open-source frameworks such as CogVideoX~\cite{yang2024cogvideox}, LTX-Video~\cite{HaCohen2024LTXVideo}, HunyuanVideo~\cite{kong2024hunyuanvideo}, and Wan~\cite{wang2025wan}, whose generation quality now rivals or even surpasses certain proprietary commercial models. These models have demonstrated promising application value and practical potential in downstream tasks~\cite{wu2024videomaker,wu2025customcrafter,li2025realcam,zheng2024cami2v,liu2023stylecrafter,ma2024follow,ma2024follow-e}.

A critical factor contributing to their success lies in the utilization of high-quality training data. Although numerous open-source models have successfully showcased the potential of video generation systems and disclosed their data processing pipelines, the collection and refinement of large-scale video-text pairs for pretraining remain exceptionally labor-intensive and resource-demanding. This has imposed constraints on open-source initiatives related to training data.

Early video datasets were primarily designed for non-generative tasks such as video understanding~\cite{Rohrbach_2015_CVPR,caba2015activitynet,sanabria2018how2,xu2016msr}, retrieval~\cite{jiang2019svd}, visual question answering (VQA)~\cite{tapaswi2016movieqa,kim2017deepstory}, or specific video generation downstream tasks~\cite{zheng2025realcam}. Koala-36M~\cite{wang2024koala} improved upon Panda-70M~\cite{panda70m} by optimizing video transition detection algorithm, implementing superior data filtering methods, and introducing structured caption system, thereby constructing a large-scale pretraining dataset suitable for video generation models. However, despite employing a quality control model with threshold-based filtering to eliminate low-quality videos, Koala-36M still exhibits limitations in visual quality. Particularly for model post-training or quality-tuning stage that demand extremely high-quality datasets, Koala-36M fails to meet the stringent quality requirements.

Both HunyuanVideo and Wan's data processing pipelines rely heavily on expert manual annotation for selecting high-quality data for supervised fine-tuning (SFT). Even after preliminary machine-based quality filtering, the retention rate remains remarkably low. Our investigation reveals that UGC platforms host numerous content creators producing high-quality videos. By manually curating creators or videos and applying simple data processing, we can efficiently collect substantial quantities of high-quality video clips.

We present Tiger200K, a manually curated dataset of high-quality video clips sourced from UGC platforms, featuring bilingual (Chinese-English) fine-grained captions generated using state-of-the-art open-source visual large language models. This dataset aims to provide high-quality video-text pairs for post-training and quality-tuning of video generation models, thereby accelerating community research on video-related generative AI.

\section{Data Processing Pipeline}

\label{sec:pipeline}
\begin{figure*}[h!]
\centering
\includegraphics[width= 0.9\linewidth]{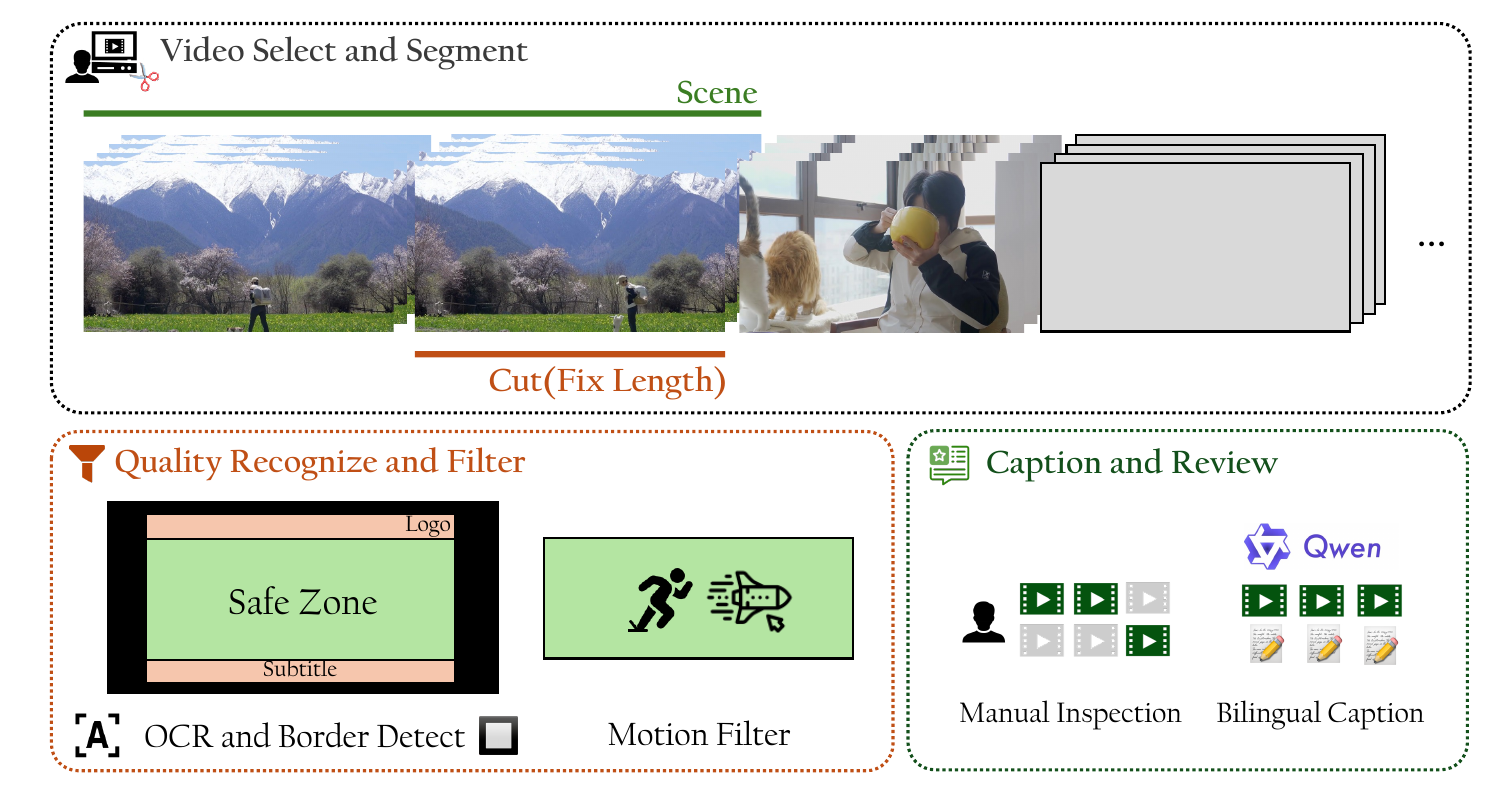}
   \caption{The pipeline of data construction. First, the selected video will segment by scene and further subdivided into cuts of fix length. Then, methods such as OCR, border detection, and optical flow estimation are used for quality filtering. Finally, manual review was carried out and bilingual fine-grained caption was performed using VLM.}
\label{fig:pipeline}
\end{figure*}

Our data processing pipeline is illustrated in Figure \ref{fig:pipeline}. The pipeline begins with manual curation of videos or video collections from UGC platforms. We employ TransNetV2~\cite{soucek2024transnet} for scene boundary detection and segmentation, followed by OCR processing and black border detection on each frame to identify safe zone for video cuts. After applying motion filtering and manual quality reviewing, we generate captions for every cuts using the state-of-the-art (SOTA) visual large language models. During both the manual curation and quality filtering stages, our primary focus remains on visual and aesthetic quality assessment.

\subsection{Source}

\begin{figure*}[h!]
\centering
\includegraphics[width= 0.9\linewidth]{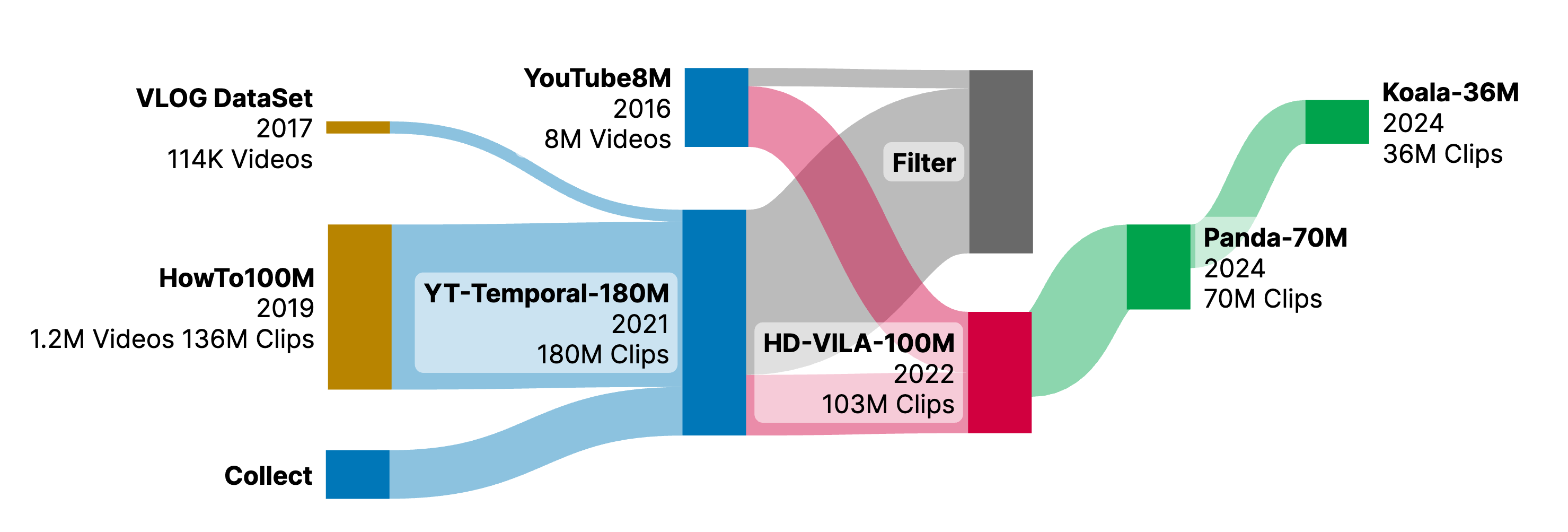}
   \caption{The inheritance relationship among Koala-36m and other datasets. Most are collected from the early Youtube content. The quantities in the figure are relative.}
\label{fig:koala36m}
\end{figure*}

As illustrated in Figure \ref{fig:koala36m}, existing datasets~\cite{fouhey2018lifestyle, miech2019howto100m, zellers2021merlot, abu2016youtube8m, hdvila, panda70m, wang2024koala} predominantly source their content from early YouTube videos. With recent advancements in multimedia technology, both the quantity and quality of user-generated content (UGC) have experienced exponential growth. Our dataset expands the temporal and spatial dimensions of data sources by incorporating freshly curated content from BiliBili, China's largest UGC video platform.

The primary criteria for our video selection emphasize visual and aesthetic quality. We employ the following strategies to efficiently identify high-quality candidate videos:

\begin{enumerate}
\item Focusing on top-tier content creators: Established creators typically maintain superior production standards and aesthetic appeal.
\item Keyword-based filtering: Systematic searches using terms like high-end camera models (e.g., A7S3, FX3, GH7, R5M2), short films, and 4K HDR.
\item Leveraging recommendation algorithms: Discovering additional premium content through related recommendations and platform-wide suggestions, benefiting from advancements in recommendation systems.
\end{enumerate}

\subsection{Shot Detecting}

Our comprehensive video analysis reveals that the vast majority of video transitions employ either hard cuts or cross dissolves as their primary editing techniques. While PySceneDetect~\cite{Castellano_PySceneDetect} demonstrates high detection accuracy for hard cuts using its default parameters, it frequently fails to reliably identify cross dissolves.

\begin{figure*}[h!]
\centering
\includegraphics[width= 1\linewidth]{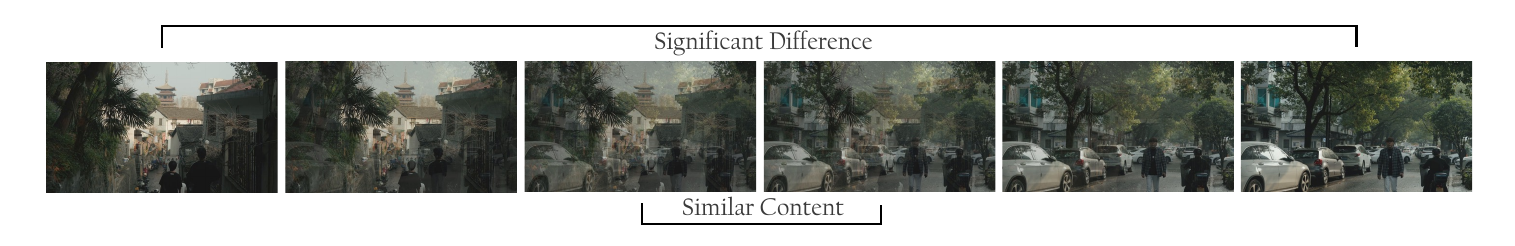}
   \caption{The content differences between the video frames of dissolve transition are relatively small, resulting in the failure of the video scene boundary detecting algorithm based on statistical differences.}
\label{fig:dissolve}
\end{figure*}

We attribute this limitation to the inherent nature of cross dissolves: the smoothing effect applied between adjacent scenes results in statistically insignificant content variations across transition frames. As illustrated in Figure \ref{fig:dissolve}, the inter-frame content differences during cross dissolve transitions exhibit little discernible variation.

To address this issue, we experimented with frame skipping techniques to amplify content dissimilarity between video clips, hypothesizing that this approach might enhance PySceneDetect's detection performance. However, our experimental results indicate that this modification yields unsatisfactory improvements in cross dissolve detection accuracy.

\begin{figure*}[h]
\centering
\includegraphics[width= 1\linewidth]{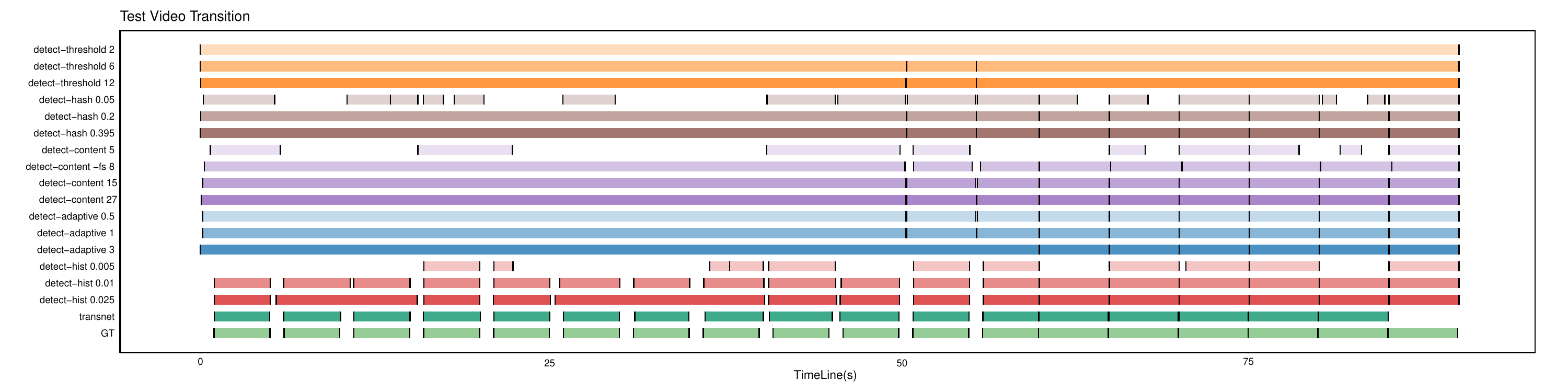}
\includegraphics[width= 1\linewidth]{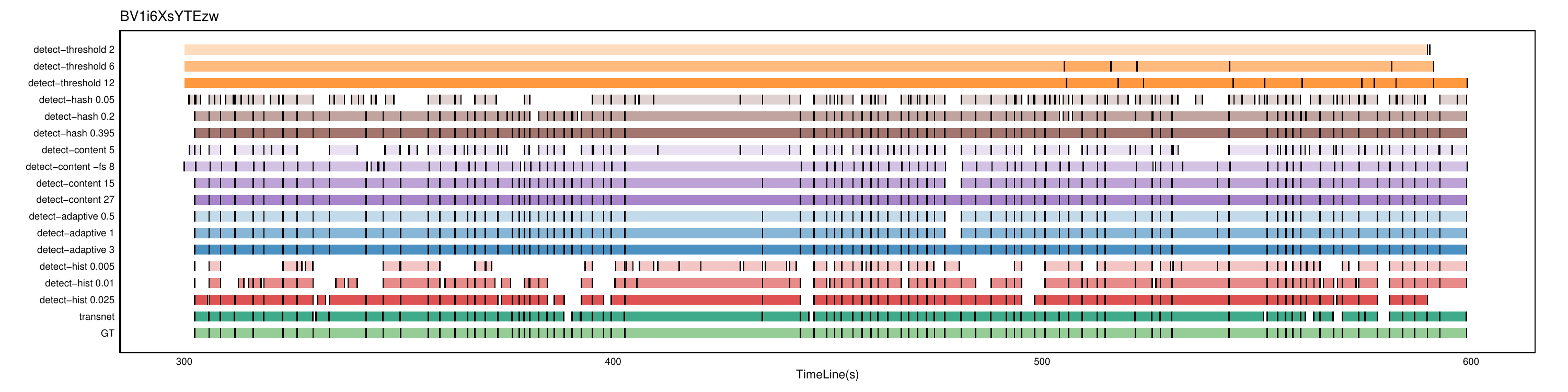}
   \caption{Comparison of the shot detecting results of PySceneDetect with different parameters and TransNetv2 on synthetic test videos (variety of transition methods) and real-world UGC videos.}
\label{fig:transition}
\end{figure*}

TransNetV2~\cite{soucek2024transnet}, a deep learning-based approach, employs a lightweight network architecture that significantly enhances shot boundary detection accuracy through convolutional operations and carefully curated training data. As demonstrated in Figure \ref{fig:transition}, our experiments show that TransNetV2 achieves consistent segmentation performance across both synthetic test videos and real-world UGC content.

We define the segments extracted by TransNetV2 as Scenes. To accommodate the requirement for temporally consistent segments in video generation model training, we discard any Scene shorter than 121 frames. For qualifying Scenes, we perform center-based segmentation to produce fixed-length (121-frame) sub-segments, which we term Cuts.

\subsection{Safe Zone}

\begin{algorithm}
\caption{Safe Zone Detect}\label{alg:safezone}
\begin{algorithmic}[1]
\Require Video $V$, OCR confidence $\tau$, border threshold $T$, text area threshold $\epsilon$, subtitle location $\alpha$
\Ensure Safe Zone $(\mathcal{X}_{1}, \mathcal{Y}_{1}, \mathcal{X}_{2}, \mathcal{Y}_{2})$

\State Initialize $\mathcal{O}_{top} \gets 0$, $\mathcal{O}_{bottom} \gets 1$ \Comment{OCR boundaries}
\State Initialize $\mathcal{B}_t \gets H$, $\mathcal{B}_b \gets 0$, $\mathcal{B}_l \gets W$, $\mathcal{B}_r \gets 0$ \Comment{Border boundaries}
\State $\Phi \gets 0$ \Comment{Middle region text area accumulator}

\For{frame $f_i \in V$}
    \State {\color{blue} \textproc{OCR Processing:}}
    \State $\text{ocr}_i \gets \text{PaddleOCR}(f_i, \tau)$ \Comment{Get bounding boxes with confidence $\geq\tau$}
    \State $\beta_{top} \gets 0$, $\beta_{bottom} \gets 1$, $\phi_i \gets 0$
    
    \For{box $b_j \in \text{ocr}_i$}
        \State $(x_1, y_1, x_2, y_2) \gets \text{Normalize}(b_j)$ \Comment{Normalized coordinates}
        \If{$y_2 \leq \alpha$} \Comment{Subtitle region}
            \State $\beta_{top} \gets \max(\beta_{top}, y_2)$
        \ElsIf{$y_1 \geq 1-\alpha$}
            \State $\beta_{bottom} \gets \min(\beta_{bottom}, y_1)$
        \Else \Comment{Middle region}
            \State $\phi_i \gets \phi_i + (y_2-y_1)(x_2-x_1)$
        \EndIf
    \EndFor
    
    \State $\mathcal{O}_{top} \gets \max(\mathcal{O}_{top}, \beta_{top})$, $\mathcal{O}_{bottom} \gets \min(\mathcal{O}_{bottom}, \beta_{bottom})$
    \State $\Phi \gets \Phi + \phi_i$

    \State {\color{blue} \textproc{Border Detection:}}
    \State $(t_i, b_i, l_i, r_i) \gets \text{ScanBorders}(f_i, T)$
    \State $\mathcal{B}_t \gets \min(\mathcal{B}_t, t_i/H)$, $\mathcal{B}_b \gets \max(\mathcal{B}_b, (H-b_i)/H)$
    \State $\mathcal{B}_l \gets \min(\mathcal{B}_l, l_i/W)$, $\mathcal{B}_r \gets \max(\mathcal{B}_r, (W-r_i)/W)$
\EndFor

\State {\color{teal} \textproc{Merge Boundaries:}}
\State $t_{final} \gets \max(\mathcal{O}_{top}, \mathcal{B}_t)$, $b_{final} \gets \min(\mathcal{O}_{bottom}, \mathcal{B}_b)$
\State $l_{final} \gets \mathcal{B}_l$, $r_{final} \gets \mathcal{B}_r$

\If{$\frac{\Phi}{|V| \cdot W \cdot H} > \epsilon$} \Comment{Too many text in video}
    \State \Return \text{Invalid}
\EndIf

\If{$(\mathcal{Y}_{2} - \mathcal{Y}_{1}) (\mathcal{X}_{2} - \mathcal{X}_{1}) < 0.5$} \Comment{Safe zone is too small}
    \State \Return \text{Invalid}
\EndIf

\State \Return $(\mathcal{X}_{1}, \mathcal{Y}_{1}, \mathcal{X}_{2}, \mathcal{Y}_{2})$
\end{algorithmic}
\end{algorithm}

In a typical video, most watermarks and subtitles are concentrated in fixed regions. Inspired by the film industry, we introduce the concept of a safe zone, defined as the area of the frame unaffected by watermarks, subtitles, logos, or other overlays.

To determine the safe zone, we employ PaddleOCR~\cite{li2022pp} to detect both Chinese and English text and a search algorithm automatically identifies black aera in all directions of the video frame. For critical regions (the top and bottom 20\% of the frame), the safe zone must completely exclude any detected text. For the central areas, we first filter out extremely small bounding boxes (bboxes) and then calculate the area ratio of text regions across all frames. A threshold is set to discard videos with excessive text coverage.

For black broader detection, we scan each frame row-wise from the top and bottom and column-wise from the left and right. The scanning terminates when the proportion of black pixels falls below a predefined threshold, thereby determining the boundaries of potential black broader. To ensure robustness, only regions consistently identified as black broader across all frames within a video clip are retained as the final detection result.

The final safe zone is determined by the intersection of the OCR-detected text-free area and the black broader detection results. If the safe zone area is less than 50\% of the original frame size, the video clip will be dropped. The complete safe zone computation process is illustrated in Algorithm \ref{alg:safezone}.

\subsection{Quality Filter}

Video clips with reasonable motion can help the model better understand spatiotemporal relationships. To obtain video clips with valid motion, we employ OpenCV to compute Farneback dense optical flow and filter out frames below a predefined threshold, thereby excluding clips with minimal subject movement or camera motion.

Additionally, we conduct manual random sampling of videos to verify the accuracy of the aforementioned procedures and ensure overall video quality.

\subsection{Video Caption}

DALL-E3~\cite{dalle3} has demonstrated that training models with detailed textual descriptions can significantly enhance the prompt-following capability and output quality of visual generation models. Leveraging recent advancements in large vision-language models, we employed the state-of-the-art Qwen2.5-VL~\cite{bai2025qwen2} to caption video in both Chinese and English, successfully obtaining dense captions for each video in our dataset.

\section{Results and Statistics}
We finally obtained 85k scene segmentation and 170k video clips from 4151 selected videos through automated data processing pipelines presented in \ref{sec:pipeline} and manual filtering, and annotated the safe zone cropping region and bilingual caption. 
We randomly selected some clips for visualization as shown in Figure \ref{fig:visualization}, these clips demonstrate
enhanced visual and aesthetic quality.

\begin{figure*}[h!]
\centering
\includegraphics[width= 1\linewidth]{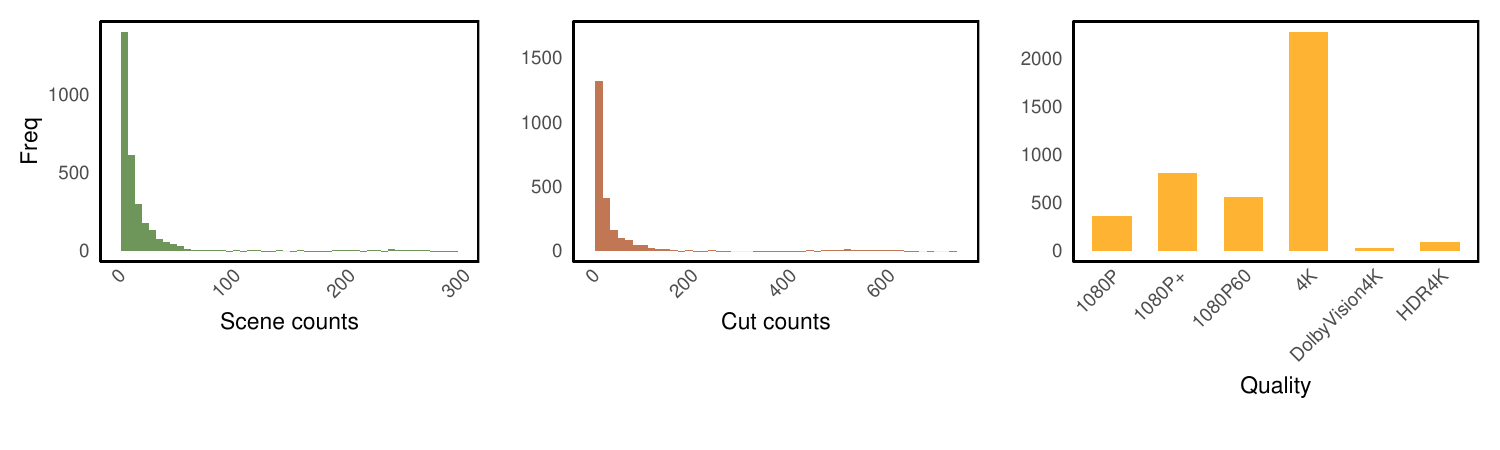}
   \caption{Statistical results of video level. Videos with 4K and 1080P resolutions each account for approximately half. The vast majority of videos can only extract a small number of clips due to their short duration and strict filtering.}
\label{fig:statistics_video}
\end{figure*}

As illustrated in Figure \ref{fig:statistics_video}, only scenes exceeding 121 frames are retained and segmented into cuts, combined with multi-stage quality filter pipeline, resulting in an average yield of fewer than 20 scenes or 40 clips per source video. Furthermore, benefiting from advancements in video platforms and the widespread adoption of 4K recording and playback devices, half of the source videos in Tiger200K have a resolution of 4K or above. This positions the dataset as a foundational resource for future research in such as high-resolution video generation or video super-resolution, providing high-quality dataset materials for these emerging computational tasks.

\begin{figure*}[h!]
\centering
\includegraphics[width= 1\linewidth]{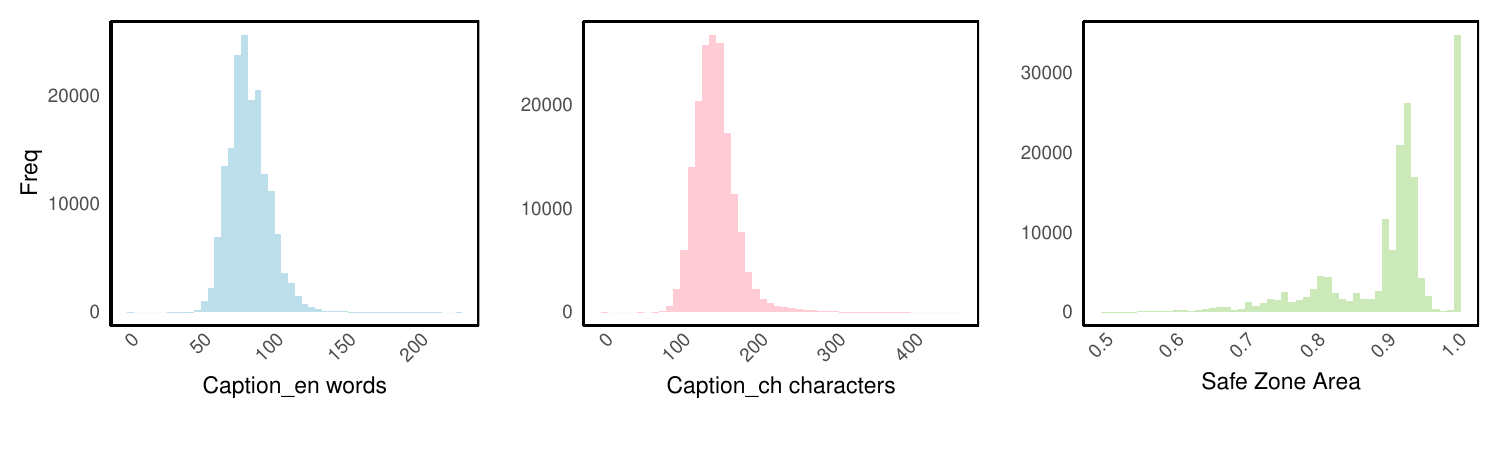}
   \caption{Statistical results of annotation level. The length of bilingual annotations is mostly controlled within a reasonable range. The retention area of safe zone for the vast majority of clips is also above 0.85.}
\label{fig:statistics_annotation}
\end{figure*}

Regarding the final annotation outcomes, our analysis reveals that caption lengths follow a normal distribution, with both English (approximately 80 words) and Chinese (around 150 characters) variants concentrated within reasonable ranges. Moreover, the majority of videos retained over 85\% of the original area after safe-zone cropping, demonstrating the high quality of the processed data and its suitability for downstream generative tasks.

\section{Conclusion}

We present Tiger200K, a manually curated high quality video dataset collected from UGC platforms, designed to address the scarcity of high quality video data in the open-source community. Although employing a relatively straightforward processing pipeline, our dataset demonstrates strong competitiveness in visual quality, a direct result of rigorous human quality control at the input stage. We commit to ongoing dataset expansion and pipeline refinement, with periodic releases to the open-source community.


\end{document}